\documentclass[runningheads,a4paper]{llncs}

\usepackage{amssymb}
\usepackage{graphicx}
\usepackage{bm}
\usepackage{bbm}
\usepackage{cite}
\usepackage{xcolor}
\usepackage{amsmath}
\usepackage{mathtools}

\DeclarePairedDelimiter\abs{\lvert}{\rvert}%
\DeclarePairedDelimiter\norm{\lVert}{\rVert}%

\makeatletter
\let\oldabs\abs
\def\abs{\@ifstar{\oldabs}{\oldabs*}}
\let\oldnorm\norm
\def\norm{\@ifstar{\oldnorm}{\oldnorm*}}
\makeatother

\usepackage{url}
\urldef{\ruizhi}\path|ruizhi@mit.edu|
\urldef{\mailsa}\path|{alfred.hofmann, ursula.barth, ingrid.haas, frank.holzwarth,|
\urldef{\mailsb}\path|anna.kramer, leonie.kunz, christine.reiss, nicole.sator,|
\urldef{\mailsc}\path|erika.siebert-cole, peter.strasser, lncs}@springer.com|

\begin{document}

\mainmatter  

\title{Temporal Registration in In-Utero \\ Volumetric MRI Time Series}

\titlerunning{Temporal Registration in In-Utero Volumetric MRI Time Series}

%
%
\author{Ruizhi Liao$^{1}$, Esra A. Turk$^{1,2}$, Miaomiao Zhang$^{1}$, Jie Luo$^{1,2}$, \\ P. Ellen Grant$^{2}$, Elfar Adalsteinsson$^{1}$, and Polina Golland$^{1}$}

\authorrunning{R. Liao \textit{et al.}}

\institute{$^{1}$  Massachusetts Institute of Technology, Cambridge, MA, USA \\ $^{2}$ Boston Children's Hospital, Harvard Medical School, Boston, MA, USA \\ \ruizhi}

%
%

\tocauthor{R. Liao \textit{et al.}}
\toctitle{Temporal Registration in In-Utero Volumetric MRI Time Series}
\maketitle

\begin{abstract}
  We present a robust method to correct for motion and deformations in
  in-utero volumetric MRI time series. Spatio-temporal analysis of
  dynamic MRI requires robust alignment across time in the presence of
  substantial and unpredictable motion. We make a Markov assumption on
  the nature of deformations to take
  advantage of the temporal structure in the image data.  Forward
  message passing in the corresponding hidden Markov model (HMM)
  yields an estimation algorithm that only has to account for
  relatively small motion between consecutive frames. We demonstrate
  the utility of the temporal model by showing that its use improves
  the accuracy of the segmentation propagation through temporal
  registration. Our results suggest that the proposed model captures
  accurately the temporal dynamics of deformations in in-utero MRI
  time series.  
\end{abstract}

\section{Introduction}

In this paper, we present a robust method for image registration in
temporal series of in-utero blood oxygenation level
dependent~(BOLD)~MRI. BOLD MRI is a promising imaging tool for
studying functional dynamics of the placenta and fetal brain
\cite{schopf2012watching, sorensen2013changes, jie2016ismrm}. It has
been shown that changes in fetal and placental oxygenation levels with
maternal hyperoxygenation can be used to detect and characterize
placental dysfunction, and therefore hold promise for monitoring
maternal and fetal well-being \cite{aimot2013vivo}. Investigating
hemodynamics of the placenta and fetal organs necessitates robust
estimation of correspondences and motion correction across different
volumes in the dynamic MRI series. Temporal MRI data suffers from
serious motion artifacts due to maternal respiration, unpredictable
fetal movements and signal
non-uniformities~\cite{studholme2011mapping}, as illustrated in Fig.~1. Our approach
exploits the temporal nature of the data to achieve robust
registration in this challenging setup.

\begin{figure}[h]
\centerline{
\hfill
\includegraphics[width=4in]{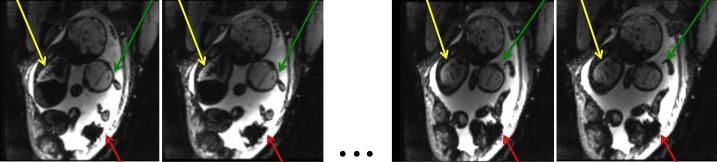}
\hfill
}
\centerline{
\hfill
\begin{minipage}[t]{0.8in}
\centering
$J_{1}$
\end{minipage} 
\hskip0.1in
\begin{minipage}[t]{0.8in}
\centering
$J_{2}$
\end{minipage} 
\hskip0.45in
\begin{minipage}[t]{0.8in}
\centering
$J_{74}$
\end{minipage} 
\hskip0.1in
\begin{minipage}[t]{0.8in}
\centering
$J_{75}$
\end{minipage} 
\hfill
}
\caption{Example twin pregnancy case from the study. The same
  cross-section from frames~$J_1$, $J_2$, $J_{74}$, and $J_{75}$ is
  shown. Arrows indicate areas of substantial motion of the placenta
  (red), fetal head (green), and fetal body (yellow). }
\vskip-0.1in
\end{figure}

Prior work in in-utero MRI has focused on the fetal brain and
demonstrated that rigid transformations capture brain motion
accurately \cite{ferrazzi2014resting, you2015robust}. The rigid model,
however, fails to fully account for movement and deformation of the
placenta.  Recently, B-spline transformations have been employed for
tracking of regions-of-interest~(ROIs) in placental images by
registering all volumes to a reference frame \cite{turk2015magnetic}.
This approach ignores the temporal nature of the data and yields a
substantial number of outlier volumes that fail registration due to
significant motion. In this paper, we demonstrate that a temporal model of movement
improves the quality of alignment.

Beyond the specific application to in-utero MRI time series, the
problem of temporal alignment has been investigated in
longitudinal~\cite{reuter2011avoiding,durrleman2013toward},
cardiac~\cite{chandrashekara2003construction,
  sundar2009biomechanically, park1996deformable, metz2011nonrigid} and
lung imaging~\cite{mcclelland2013respiratory, rietzel2006deformable,
reinhardt2008registration,metz2011nonrigid,
  singh2015hierarchical}.  Longitudinal studies often involve subtle
changes, and the algorithms are fine-tuned to detect small
deformations~\cite{reuter2011avoiding}.  Both cardiac and lung motion
patterns are somewhat regular and smooth across time, and lend
themselves to biomechanical
modeling~\cite{sundar2009biomechanically,mcclelland2013respiratory,
  park1996deformable}. In contrast to these applications, in-utero
volumetric MRI time series contain a combination of subtle non-rigid
deformation of the placenta and large-scale unpredictable motion of
the fetus.  At the same time, consecutive frames in a series are quite
close to each other in time, which is the property we exploit in our modeling.

Existing methods for temporal registration in image series can be
categorized into three distinct groups. The first group applies a
variant of groupwise registration to this problem.  This approach
relies on a group template -- estimated or selected from the image set
-- that yields acceptable registration results for all frames
\cite{rietzel2006deformable, metz2011nonrigid, durrleman2013toward,
  singh2015hierarchical}. Unfortunately, large motion present in
in-utero MRI makes some frames to be substantially different from the
template, leading to registration failures. The second group aligns
consecutive frames and concatenates resulting deformations to estimate
alignment of all frames in the series \cite{
  reinhardt2008registration}. In application to long image series
(BOLD MRI series contain hundreds of volumes), this approach leads to
substantial errors after several concatenation steps.  The third
approach formulates the objective function in terms of pairwise
differences between consecutive frames, leading to algorithms that
perform pairwise registration of consecutive frames iteratively until
the entire series comes into alignment \cite{metz2011nonrigid}. 
Our method is also related to filtering approaches in respiratory
motion modeling~\cite{mcclelland2013respiratory}. Since in-utero
motion is much more complex than respiratory motion, we do not attempt
to explicitly model motion but rather capture it through deformations
of the latent template image.

In this paper, we construct the so called filtered estimates of the
deformations by making a Markov assumption on the temporal series. We
derive a sequential procedure to determine the non-rigid
transformation of the template to each frame in the series. This work
represents a first step towards efficient and robust temporal
alignment in this challenging novel application, and provides a
flexible framework that can be augmented in the future with clinically
relevant estimation of MRI intensity dynamics in organs of interest.
We demonstrate the method on real in-utero MRI time series, and report robust improvements in alignment of placenta, fetal brains, and fetal livers.


\section{HMM and Filtered Estimates}
\label{sec:hmm}

In this section, we briefly review inference in HMMs
\cite{baum1966statistical, bishop2006pattern}, and introduce our
notation in the context of temporal registration. We assume existence
of a latent~(hidden) state whose temporal dynamics is governed by a
Markov structure, i.e., the future state depends on the history only
through the current state. In our application, we assume that template
$I$ deforms at each time point to describe the anatomical arrangement
at that time. Deformation $\varphi_n$ defines the latent state at time
$n\in\{1, ..., N\}$, where $N$ is the number of images in the series.
The observed image $J_n$ at time $n$ is generated by applying
$\varphi_n$ to template~$I$ independently of all other time points.

We aim to estimate the latent variables~$\{\varphi_n\}$ from
observations~$\{J_n\}$. Formally, we construct and then maximize
posterior distribution $p(\varphi_n|{J_{1:{n}}}; I)$, where we use
$J_{k:m}$ to denote sub-series~$\{J_k, J_{k+1}, ..., J_m\}$ of the
volumetric time series~$\{J_1, J_2, ..., J_N\}$. This distribution,
referred to as a filtered estimate of the state, can be efficiently
constructed using forward message passing \cite{bishop2006pattern,
  baum1966statistical}, also known as sequential estimation. The
message $m_{{(n-1)}\to{(n)}}(\varphi_n)$ from node $n-1$ to node $n$
is determined through forward recursion that integrates a previous
message~$m_{(n-2)\to{(n-1)}}(\varphi_{n-1})$ with the data likelihood
$p(J_n|\varphi_{n}; I)$ and dynamics~$p({\varphi_n}|\varphi_{n-1})$:
\begin{align}
m_{{(n-1)}\to{(n)}}(\varphi_n) &\triangleq p({\varphi_n}|J_{1:n}; I)  
\propto p(J_n|\varphi_{n}; I) \, p({\varphi_n}|J_{1:{n-1}}; I) \\
&= p(J_n|\varphi_{n}; I) 
\!\! \int\limits_{{{\varphi}_{{n-1}}}} \!\!\! p({\varphi_n}|\varphi_{n-1}) \,
m_{(n-2)\to{(n-1)}}(\varphi_{n-1}) \, d{\varphi_{n-1}},
\label{forwardMessagePassing}
\end{align}
where $m_{0\to1}(\varphi_1) = p(\varphi_1)$ and $n=\{1, ..., N\}$. The forward pass produces the posterior distribution $p(\varphi_n|J_{1:n}; I)$ for each time point $n$ in the number of steps that is linear with $n$. Similarly, efficient backward pass enables computation of posterior distribution $p(\varphi_n|J_{1:N}; I)$ based on all data, often referred to as smoothing. In this paper, we investigate advantages of the temporal model in the context of filtering and leave the development of a smoothing algorithm for future work.

\section{Modeling Temporal Deformations using HMM}
\label{sec:method}

The likelihood term $p(J_n|\varphi_n; I)$ in
Eq.~(\ref{forwardMessagePassing}) is determined by the model of image
noise:
\begin{equation}
\label{imageMatching}
p(J_n|\varphi_n; I) \propto \exp\left(-\text{Dist}\left(J_n , 
I \left(\varphi^{-1}_n\right)\right)\right),
\end{equation}
where $\text{Dist}(\cdot, \cdot)$ is a measure of dissimilarity
between images. The transition probability
$p({\varphi_n}|\varphi_{n-1})$ encourages temporal and spatial
smoothness:
\begin{equation}
  p({\varphi_n}|\varphi_{n-1}) \propto \exp\left(-\lambda_1 \text{Reg}{(\varphi_n)}-
    \lambda_2 {\norm*{\varphi_n\circ\varphi_{n-1}^{-1}}}^2\right),
\end{equation}
where $\text{Reg}{(\cdot)}$ is the regularization term that encourages
spatial smoothness of the deformation, $\norm*{\cdot}$ is the
appropriate norm that encourages $\varphi_n$ to be close
to~$\varphi_{n-1}$, and $\lambda_1$ and $\lambda_2$ are regularization
parameters.

Since the integration over all possible deformation fields is
intractable, we resort to a commonly used approximation of evaluating
the point estimate that maximizes the integrand. In particular, if
$\varphi_{n-1}^{*}$ is the best deformation estimated by the method
for time point $n-1$, the message passing can be viewed as passing the
optimal deformation $\varphi_{n-1}^{*}$ to node $n$:
\begin{align}
  m_{({n-1})\to({n})}(\varphi_n) & \propto p(J_n|\varphi_{n}; I) 
\!\!\int\limits_{\varphi_{n-1}}  \!\!\! p({\varphi_n}|\varphi_{n-1}) \,
\mathbbm{1}\{\varphi_{n-1}=\varphi_{n-1}^{*}\} \, d{\varphi_{n-1}} \\
\label{messagePassing}
&= p(J_{n}|\varphi_{n}; I) \, p({\varphi_{n}}|\varphi_{n-1}^{*}),
\end{align}
and the estimate for time point $n$ is recursively estimated as
\begin{equation}
\varphi_{n}^{*}=\arg\max_{\varphi_{n}}\ p(J_{n}|\varphi_{n}; I)
\, p({\varphi_{n}}|\varphi_{n-1}^{*}).
\label{max:def}
\end{equation}
This estimate is then used to determine $\varphi_{n+1}^{*}$, and so on
until we reach the end of the series.

\section{Implementation} 
\label{sec:impl}

In this work, we choose the first image $J_1$ as the reference
template $I$ and focus on exploring the advantages of the Markov
structure. The model can be readily augmented to include a model of a
bias field and a latent reference template that is estimated jointly
with the deformations, similar to prior work in groupwise registration
\cite{metz2011nonrigid, durrleman2013toward,
  singh2015hierarchical,rietzel2006deformable}.  We manipulate
Eq.~(\ref{max:def}) to obtain
\begin{align}
\label{messagePassing1}
\varphi_{n}^{*} & = \arg\max_{\varphi_{n}}\ p(J_n|\varphi_n; I) \, 
p(\varphi_n|\varphi_{n-1}^{*})  \\ 
\label{messagePassing2}
& = \arg\min_{\varphi_{n}}\ \text{Dist}\left(J_n ,
  I\left(\varphi^{-1}_n\right)\right)+ \lambda_1
\text{Reg}{(\varphi_n)}+ \lambda_2
{\norm*{\varphi_n\circ{(\varphi_{n-1}^{*})}^{-1}}}^2,
\end{align}
and observe that this optimization problem reduces to pairwise image
registration of the template $I$ and the observed image $J_n$. The
algorithm proceeds as follow. Given the estimate $\varphi_{n-1}^{*}$
of the template deformation to represent image~$J_{n-1}$, we apply the
registration algorithm to $I$ and $J_n$ while using
$\varphi_{n-1}^{*}$ as an initialization, resulting in the
estimate~$\varphi_{n}^{*}$.


We implemented our method using symmetric diffeomorphic registration
with cross-correlation \cite{avants2008symmetric}. Diffeomorphic
registration ensures that the estimated deformation is differentiable
bijective with differentiable inverse. We employ cross-correlation to
define the measure of image dissimilarity $\text{Dist}(\cdot, \cdot)$, because cross-correlation adapts naturally to the image data with
signal non-uniformities. We set the size of the local window for
computing cross-correlation to be 5 voxels. We use the
state-of-the-art implementation provided in the ANTS software package
\cite{avants2008symmetric}.  Following the common practice in the
field, ANTS implements spatial regularization via Gaussian smoothing.


\section{Experiments and Results}
\label{sec:results}

\paragraph{\bf Data } Ten pregnant women were consented and scanned on
a 3T Skyra Siemens scanner (single-shot GRE-EPI, $3 \times 3
{\text{mm}}^2$ in-plane resolution, $3 \text{mm}$ slice thickness,
interleaved slice acquisition, TR$=5.8-8 \text{s}$, TE$=32-36
\text{ms}$, FA$=90^o$) using 18-channel body and 12-channel spine
receive arrays.  Each series contains around 300 volumes. To eliminate
the effects of slice interleaving, we resampled odd and even slices of
each volume onto a common isotropic~$3$mm$^3$ image grid.  This study
included three singleton pregnancies, six twin pregnancies, and one
triplet pregnancy, between 28 and 37 weeks of gestational age. A
hyperoxia task paradigm was used during the scans, comprising three
consecutive ten-minute episodes of initial normoxic episode~($21\%
\text{O}_2$), hyperoxic episode~($100\% \text{O}_2$), and a final
normoxic episode~($21\% \text{O}_2$).  To enable quantitative
evaluation, we manually delineated the placentae (total of 10), fetal
brains (total of 18), and fetal livers (total of 18), in the reference
template $I=J_1$ and in five additional randomly chosen volumes in
each series.

\paragraph{\bf Experiments } To evaluate the advantages of the
temporal model, we compare it to a variant of our algorithm that does
not assume temporal structure and instead aligns each image in the
series to the reference frame using the same registration algorithm
used by our method. Algorithmically, this corresponds to setting $\lambda_2$ in Eq.~(\ref{messagePassing2}) to be $0$, and initializing
the registration step with an identity
transformation instead of the previously estimated
transformation~$\varphi_{n-1}^*$. To quantify the accuracy of the
alignment, we transform the manual segmentations in the reference
template to the five segmented frames in each series using the
estimated deformations. We employed Dice
coefficient~\cite{dice1945measures} to quantify volume overlap between
the transferred and the manual segmentations. In our application, the goal is to study average temporal signals for each ROI,
and therefore delineation of an ROI provides an appropriate evaluation target.

\begin{figure}[t]
\centerline{
\hfill
\hskip0.2in
Case 1
\hskip2.1in
Case 2
\hskip0.2in
\hfill
}
\centerline{
\includegraphics[width=2.2in]{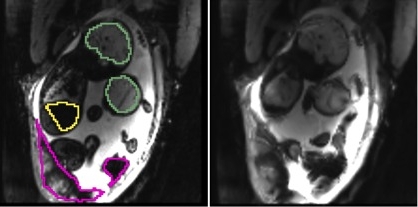}
\hfill
\includegraphics[width=2.2in]{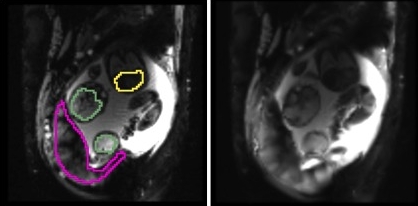}
}
\centerline{
\hfill
\begin{minipage}[t]{0.9in}
\centering
$J_1$, manual 
\end{minipage} 
\hskip0.15in
\begin{minipage}[t]{0.9in}
\centering
$J_1(\varphi_{75}^{-1})$
\end{minipage} 
\hskip0.45in
\begin{minipage}[t]{0.9in}
\centering
$J_1$, manual 
\end{minipage} 
\hskip0.15in
\begin{minipage}[t]{0.9in}
\centering
$J_1(\varphi_{75}^{-1})$
\end{minipage} 
\hfill
}
\vskip0.1in
\centerline{
\includegraphics[width=2.2in]{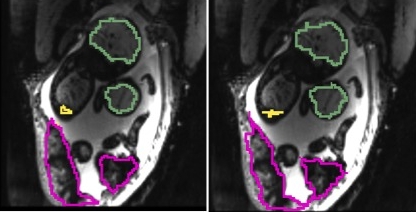}
\hfill
\includegraphics[width=2.2in]{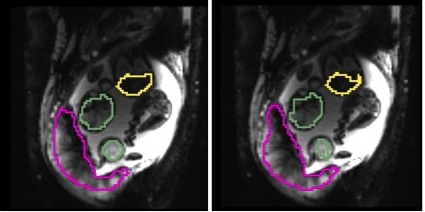}
}
\centerline{
\hfill
\begin{minipage}[t]{0.9in}
\centering
$J_{75}$, manual 
\end{minipage} 
\hskip0.15in
\begin{minipage}[t]{0.9in}
\centering
$J_{75}$, automatic
\end{minipage} 
\hskip0.45in
\begin{minipage}[t]{0.9in}
\centering
$J_{75}$, manual 
\end{minipage} 
\hskip0.15in
\begin{minipage}[t]{0.9in}
\centering
$J_{75}$, automatic
\end{minipage} 
\hfill
}
\caption{Two example cases from the study. For each case, we display
  the reference frame~$J_1$ with manual segmentations, the reference
  frame~$J_1(\varphi_{75}^{-1})$ transformed into the coordinate
  system of frame $J_{75}$, frame~$J_{75}$ with manual segmentations,
  and frame~$J_{75}$ with segmentations transferred from the reference
  frame~$J_1$ via $\varphi_{75}$. Both cases are twin pregnancies.
  Segmentations of the placentae (pink), fetal brains (green), and
  fetal livers (yellow) are shown. Two-dimensional cross-sections are
  used for visualization purposes only; all computations are performed
  in 3D.  }
\end{figure}


\begin{figure}[h]
\hskip-0.2in
\begin{minipage}{2.1in} 
\centerline{
\hfill
\includegraphics[height=1.9in]{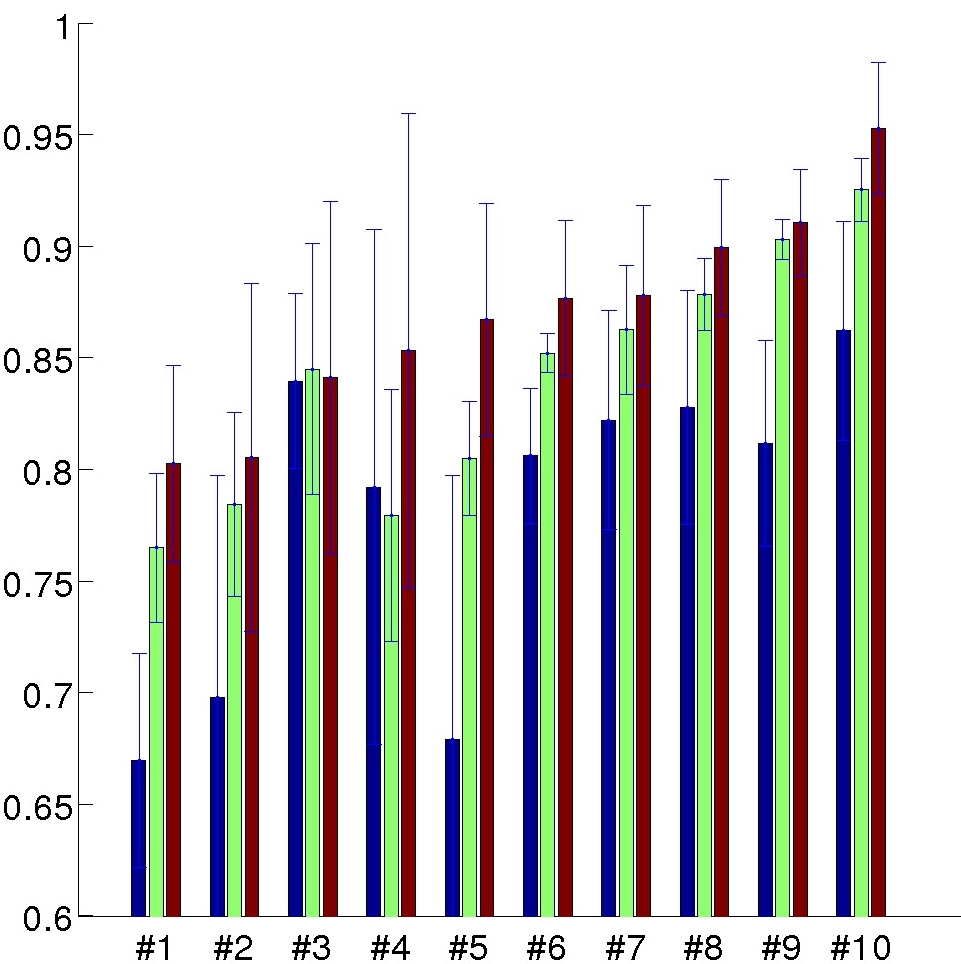}
\hfill
}
\centerline{
\hfill
(a) Placentae
\hfill
}
\end{minipage}
\hskip0.3in
\begin{minipage}{2.7in} 
\centerline{
\hfill
\includegraphics[height=1.9in]{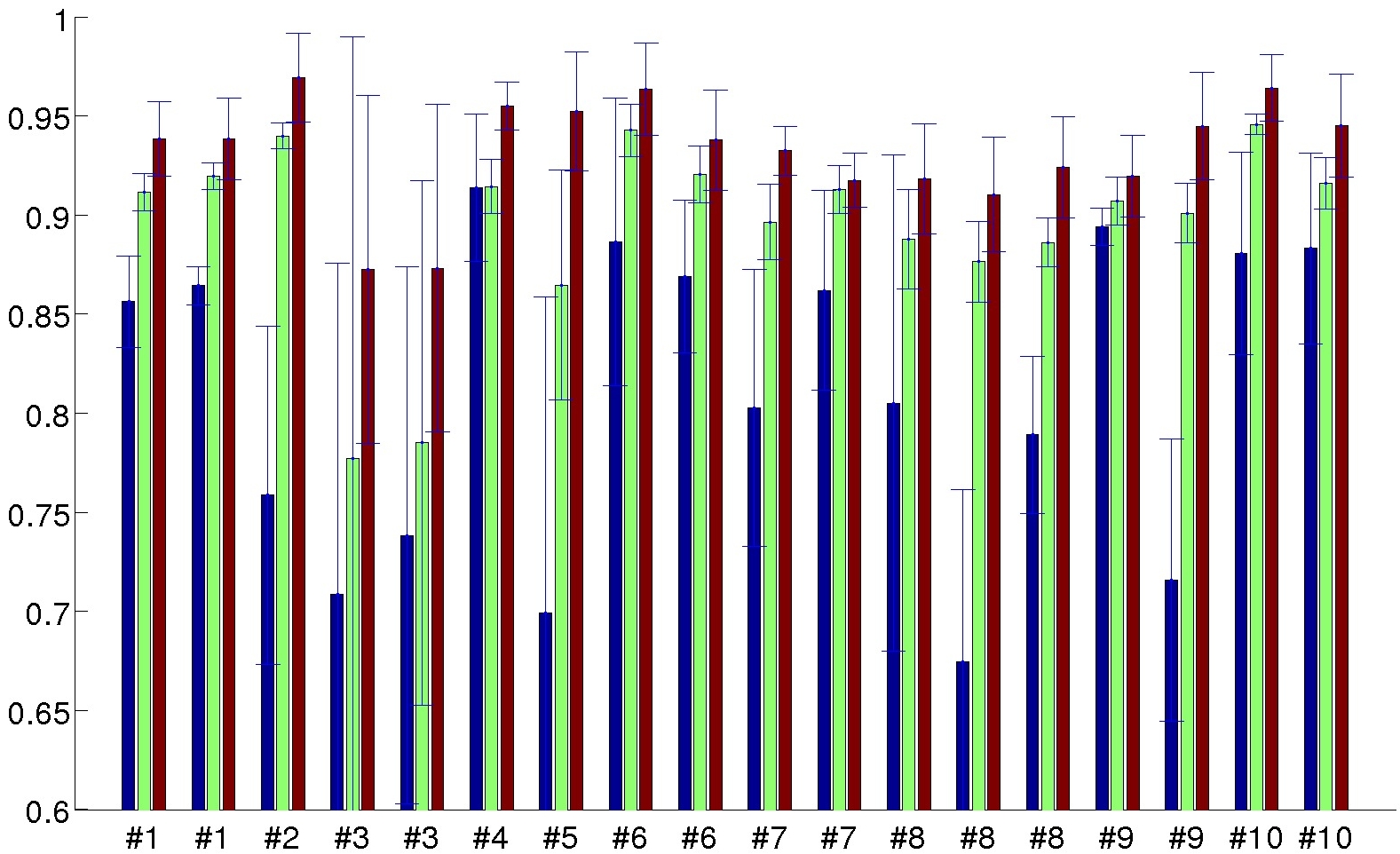}
\hfill
}
\centerline{
\hfill
(b) Fetal brains
\hfill
}
\end{minipage}

\hskip-0.2in
\begin{minipage}{2.1in} 
  \caption{Volume overlap between transferred and manual
    segmentations: (a)~placentae, (b)~fetal brains, and (c)~fetal
    livers. The cases in the study are reported in the increasing
    order of placental volume overlap for our method. Duplicate case
    numbers correspond to twin and triplet pregnancies.  Statistics
    are reported for our method (red), pairwise registration to the
    template frame (green), and no alignment (blue).}
  \vskip0.2in
\end{minipage} 
\hskip0.3in
\begin{minipage}{2.7in} 
\centerline{
\hfill
\includegraphics[height=1.9in]{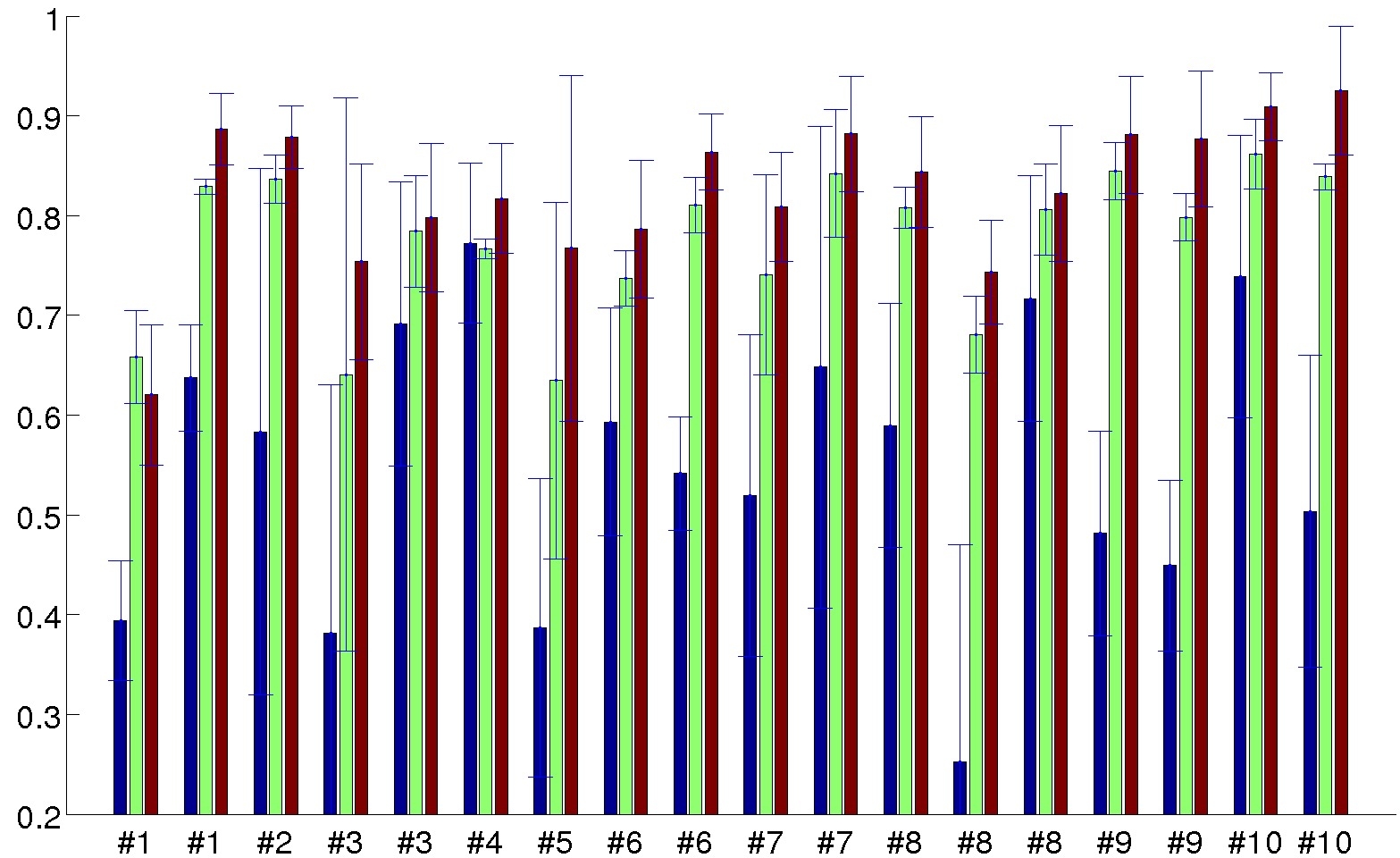}
\hfill
}
\centerline{
\hfill
{(c) Fetal livers}
\hfill
}
\end{minipage} 
\end{figure}

\paragraph{\bf Experimental Results } 
Fig.~2 illustrates results for two example cases from the study.  We
observe that the reference frame was warped accurately by the
algorithm to represent a frame in the series that is substantially
different in the regions of the placenta and the fetal liver. The
delineations achieved by transferring manual segmentations from the
reference frame to the coordinate system of the current frame
($J_{75}$ in the figure) are in good alignment with the manual
segmentations for the current frame. Fig.~3 reports volume overlap
statistics for the placentae, fetal brains, and fetal livers, for each
case in the study.  We observe that temporal alignment improves volume
overlap in important ROIs and offers consistent improvement for all
cases over pairwise registration to the reference frame. We also note
that temporal alignment offers particularly substantial gains in cases
with a lot of motion, i.e., low original volume overlap.

\section{Conclusions}

We presented a HMM-based registration method to align images in
in-utero volumetric MRI time series. Forward message passing
incorporates the temporal model of motion into the estimation
procedure. The filtered estimates are therefore based on not only the
present volume frame and the template, but also on the previous frames
in the series. The experimental results demonstrate the promise of our
approach in a novel, challenging application of in-utero BOLD MRI
analysis.  Future work will focus on obtaining robust estimates of the
MRI signal time courses by augmenting the method with a backward pass and a model of ROI-specific intensity changes.

\subsubsection*{Acknowledgments.} This work was supported in part by NIH NIBIB NAC P41EB015902, NIH NICHD U01HD087211, NIH NIBIB R01EB017337, Wistron Corporation, and Merrill Lynch Fellowship.

\bibliographystyle{splncs}
\bibliography{paper1168}

\end{document}